\documentclass[letterpaper]{article} 
\usepackage{aaai2026}  
\usepackage{times}  
\usepackage{helvet}  
\usepackage{courier}  
\usepackage[hyphens]{url}  
\usepackage{graphicx} 
\usepackage{natbib}  
\usepackage{caption} 
\usepackage{algorithm}
\usepackage{algorithmic}

\usepackage{newfloat}
\usepackage{listings}
\usepackage{colortbl}
\usepackage{color}
\usepackage{placeins}
\usepackage{arydshln}
\usepackage{multirow}
\usepackage{algorithmic}
\usepackage{algorithm}
\usepackage{booktabs}
\usepackage{amssymb}
\usepackage[capitalise]{cleveref}

\DeclareCaptionStyle{ruled}{labelfont=normalfont,labelsep=colon,strut=off} 
\floatstyle{ruled}
\newfloat{listing}{tb}{lst}{}
\floatname{listing}{Listing}
\nocopyright

\title{Q-CLIP: Unleashing the Power of Vision-Language Models for Video Quality Assessment through Unified Cross-Modal Adaptation}

\author{
    Yachun Mi\textsuperscript{\rm 1}, Yu Li\textsuperscript{\rm 1}, Yanting Li\textsuperscript{\rm 1}, Chen Hui\textsuperscript{\rm 1,2}, Tong Zhang\textsuperscript{\rm 1}, \\ Zhixuan Li\textsuperscript{\rm 3}, Chenyue Song\textsuperscript{\rm 1}, Wei Yang Bryan Lim\textsuperscript{\rm 3}, Shaohui Liu\textsuperscript{\rm 1*} 
}
\affiliations{
    \textsuperscript{\rm 1}Harbin Institute of Technology\\
    \textsuperscript{\rm 2}Nanjing University of Information Science and Technology\\
    \textsuperscript{\rm 3}Nanyang Technological University\\

}

\usepackage{bibentry}

\begin{document}

\maketitle

\begin{abstract}
Accurate and efficient Video Quality Assessment (VQA) has long been a key research challenge.
Current mainstream VQA methods typically improve performance by pretraining on large-scale classification datasets (e.g., ImageNet, Kinetics-400), followed by fine-tuning on VQA datasets.
However, this strategy presents two significant challenges: (1) merely transferring semantic knowledge learned from pretraining is insufficient for VQA, as video quality depends on multiple factors (e.g., semantics, distortion, motion, aesthetics); (2) pretraining on large-scale datasets demands enormous computational resources, often dozens or even hundreds of times greater than training directly on VQA datasets. Recently, Vision-Language Models (VLMs) have shown remarkable generalization capabilities across a wide range of visual tasks, and have begun to demonstrate promising potential in quality assessment. In this work, we propose Q-CLIP, the first fully VLMs-based framework for VQA. Q-CLIP enhances both visual and textual representations through a Shared Cross-Modal Adapter (SCMA), which contains only a minimal number of trainable parameters and is the only component that requires training. This design significantly reduces computational cost. In addition, we introduce a set of five learnable quality-level prompts to guide the VLMs in perceiving subtle quality variations, thereby further enhancing the model’s sensitivity to video quality.
Furthermore, we investigate the impact of different frame sampling strategies on VQA performance, and find that frame-difference-based sampling leads to better generalization performance across datasets. Extensive experiments demonstrate that Q-CLIP exhibits excellent performance on several VQA datasets.

\end{abstract}


\section{Introduction}
The widespread availability of portable filming devices has significantly lowered the cost and barriers to video production.
As a result, a massive volume of videos with substantial quality issues has been uploaded to the Internet. 
Since video quality directly affects users' Quality of Experience (QoE), effective Video Quality Assessment (VQA) methods are essential for identifying and filtering poor-quality content.
\begin{figure}[h]
	\centering
	\includegraphics[scale=0.2]{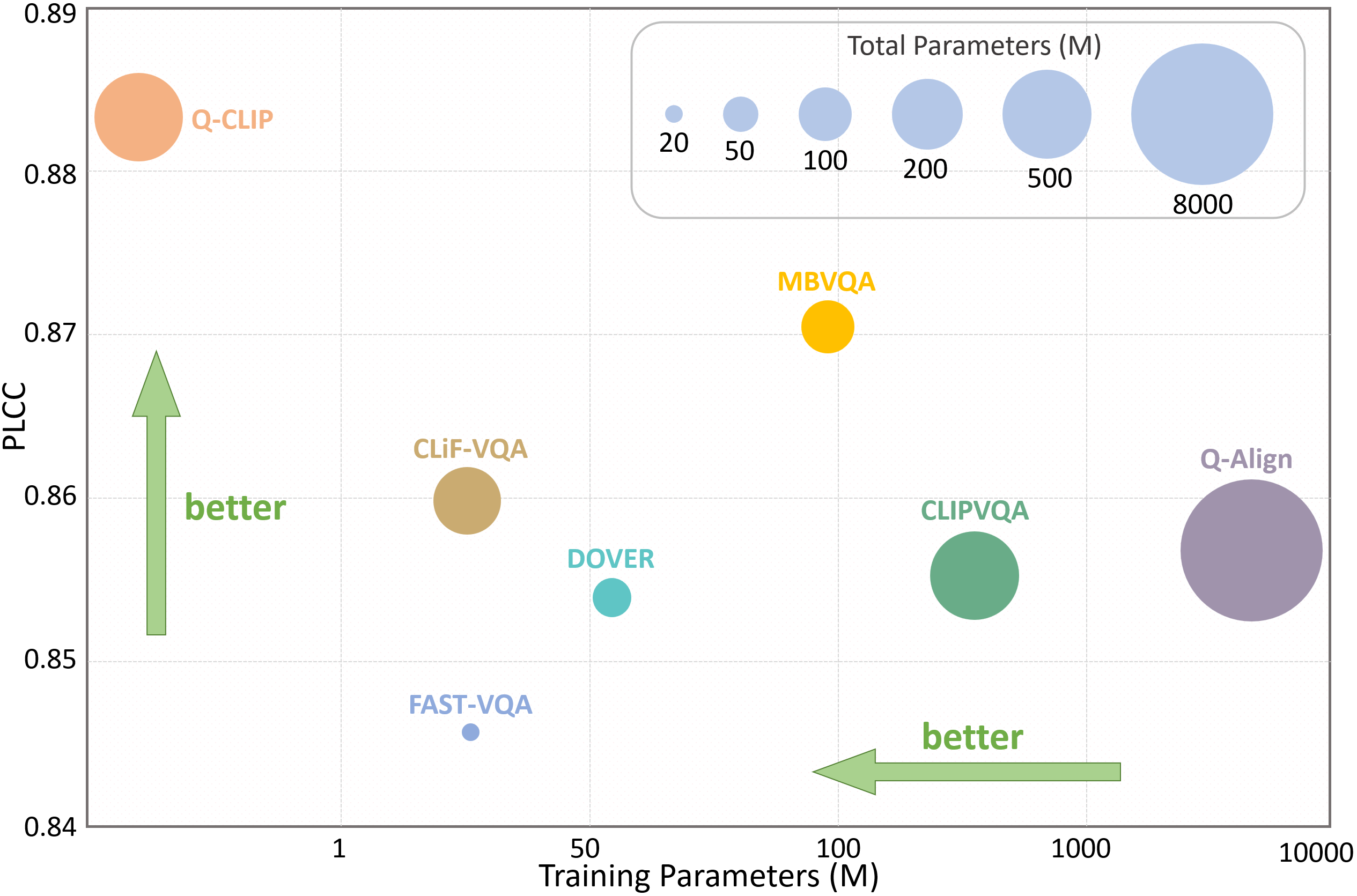}
	\caption{Comparison of Q-CLIP with leading VQA methods on LSVQ.
	Q-CLIP achieves the best performance while training only a minimal number of parameters. 
	}
	\label{fig5}
\end{figure}

Current VQA models are categorized into knowledge-driven and data-driven methods. 
Knowledge-driven methods \cite{paper33,paper11,paper10,paper8,paper7} rely on handcrafted features to characterize video quality. However, these features are often insufficient to capture the complex factors affecting quality, leading to unreliable predictions.
Benefiting from several high-quality VQA datasets obtained through subjective experiments, data-driven methods \cite{paper15,paper16,paper17,paper18,paper29,paper23,paper19} can train Deep Neural Networks (DNNs) to automatically learn complex and abstract representations of video quality, thereby outperforming approaches based on handcrafted features.
However, the high annotation cost of subjective testing leads to VQA datasets being much smaller than the large-scale datasets used in other vision tasks, which limits the full potential of deep learning models in VQA tasks.


To address the data scarcity issue, the mainstream solutions adopt a "pretraining-finetuning" paradigm: models are first pre-trained on large-scale classification datasets (e.g., ImageNet \cite{paper57}, Kinetics-400 \cite{paper65}), and then fine-tuned on VQA datasets \cite{paper1,paper2,paper3,paper16,paper6,paper107}. While this approach enhances performance, it introduces two critical limitations. 
First, classification-based pretraining focuses primarily on semantic learning, which only partially captures the perceptual aspects of video quality. 
Studies \cite{paper29,paper15,paper22,paper80,paper108} have shown that video quality depends on multiple dimensions, including semantics, distortion, motion, aesthetics, etc.,  many of which are not effectively represented by semantic classification alone.
Therefore, semantic knowledge learned from classification tasks is inherently limited in its ability to represent overall video quality.
Second, the pretraining stage demands substantially more computational resources, often by orders of magnitude compared to finetuning.
Taking FAST-VQA \cite{paper29} as an example, pretraining on Kinetics-400 is roughly 10× and 200× more expensive than finetuning on LSVQ \cite{paper16} and KoNViD-1k \cite{paper1}, respectively.

Recent advances in Vision-Language Models (VLMs) \cite{paper24,paper109,paper110,paper112,paper113} have introduced new perspectives for solving visual tasks. 
Trained on large-scale image–text pairs, these models acquire rich multimodal knowledge and demonstrate impressive generalization capabilities across domains \cite{paper114}. Unlike traditional classification-based pretraining, which primarily emphasizes semantic discrimination, VLMs inherently encode cross-modal representations that better reflect the multifaceted nature of video quality—spanning perceptual distortions (e.g., blur, noise), motion dynamics, aesthetic preferences, and semantic consistency.
Moreover, recent studies \cite{paper80,paper81,paper25} demonstrate that VLMs perform well in zero-shot quality prediction across multiple quality-related dimensions, even without task-specific supervision, highlighting their strong potential in quality perception.
These properties suggest that VLMs may serve as a promising alternative to classification-based pretraining strategies, offering a more holistic understanding of video quality while also alleviating the computational burden associated with large-scale pretraining. 

Despite their strong generalization, efficiently adapting VLMs to VQA remains challenging.
The performance of VLMs in downstream tasks is often constrained by limited intra- and cross-modal representational capacity, particularly in fine-grained perceptual tasks \cite{paper115,paper116,paper117}.
However, as a typical fine-grained perceptual task, VQA requires the model to capture subtle quality differences and rely heavily on localized visual cues, which further amplifies the challenge of transferring knowledge from VLMs. 
Moreover, fine-tuning VLMs for VQA not only incurs substantial computational costs but also risks degrading their original representational capabilities.
This study explores efficient strategies to improve VLMs' representation of quality-related factors with minimal computational overhead, thus enhancing their performance in video quality modeling.


In addition, fine-grained prompts play a crucial role in providing textual guidance for VLMs \cite{paper118,paper119,paper120,paper121,paper122}.
Existing VLMs-based quality assessment methods \cite{paper63,paper25} often rely on antonym pairs (e.g., "good" and "bad") to guide quality perception.
However, such binary prompts provide only coarse-grained supervision and may be insufficient for capturing the full aspects for describing the video quality \cite{paper80,paper81}. 
Recent studies \cite{paper105,paper123} in Large Language Models (LLMs)-based quality assessment show that mapping quality scores to a five-level scale (excellent, good, fair, poor, bad) leads to more accurate predictions. 
Building on this, we consider similar prompt strategies as a promising direction for enhancing VLMs in VQA.

Based on the above analysis, we introduce Q-CLIP, a VQA method based entirely on VLMs.
Specifically, we design a Shared Cross-Modal Adapter \textbf{(SCMA)} to enhance the representations of the visual and textual branches. 
This adapter consists of only a few fully connected layers (0.14M) and is the only component that requires training, significantly reducing computational overhead. 
As shown in Fig. \ref{fig5}, Q-CLIP achieves the best performance while training only a minimal number of parameters.
In addition, we develop a set of learnable five-level prompts (excellent, good, fair, poor, bad) to provide fine-grained textual quality descriptions as input guidance to the VLMs.
This allows us to jointly consider the similarity scores between the video and prompts of different quality levels, enabling more accurate quality prediction.
Furthermore, we investigate the impact of different frame sampling strategies on VQA performance. 
Previous works \cite{paper29,paper106} mainly adopt random or uniform sampling, with limited exploration of its impact on VQA performance.
Specifically, beyond conventional methods, we explore frame-difference-based sampling strategies to assess its potential benefits for VQA. Our findings offer new insights that may inform and inspire future research in this direction.

Our contributions can be summarized as follows:
\begin{itemize}
	\item We introduce Q-CLIP, the first VQA model fully based on VLMs. By incorporating an extremely lightweight adapter (SCMA), Q-CLIP effectively boosts the VQA capabilities of VLMs at a remarkably low training cost.
	\item We design a learnable five-level prompt mechanism to guide VLMs in perceiving subtle quality variations.
	\item This work presents a systematic study of frame sampling strategies, offering new insights and practical guidance for future research in VQA.
	\item Extensive experiments show that Q-CLIP achieves the best performance on multiple VQA datasets.
\end{itemize}

\section{Related Work}
\subsection{VQA methods}

\textbf{Knowledge-driven}
Knowledge-driven methods \cite{paper10,paper30,paper7,paper8,paper66,paper33,paper11} assess video quality by extracting handcrafted features.
For example, VIIDEO \cite{paper10} utilizes intrinsic statistical regularities of natural videos to capture anomalous information caused by distortion.
TLVQM \cite{paper8} extracts low-complexity motion features and high-complexity spatial features.
VIDEAL \cite{paper7} detects and quantifies distortions by extracting a diverse set of perceptual quality features.
However, handcrafted features struggle to capture the complex and diverse factors affecting video quality, leading to suboptimal performance.

\textbf{Data-driven} 
Data-driven methods automatically extract quality-aware features by training DNNs on high-quality VQA datasets.
For example, GST-VQA \cite{paper41} and VSFA \cite{paper15} use pretrained 2D Convolutional Neural Networks (CNNs) \cite{paper47,paper48} combined with GRU \cite{paper58} for spatiotemporal modeling, while other studies \cite{paper16,paper17,paper18,paper22,paper43,paper106} 
further incorporate 3D-CNNs \cite{paper50,paper51,paper52} to enhance spatiotemporal feature extraction.
In addition, Transformer-based VQA \cite{paper23,paper29,paper19,paper131,paper132} is gradually gaining more competitive performance.
For example, FAST-VQA \cite{paper29} and FasterVQA \cite{paper19} sample spatial-temporal grids and utilize modified Video Swin Transformers \cite{paper53}.
However, these fragment sampling strategies often neglect semantic content.
Based on this, DOVER \cite{paper64} and Zoom-VQA \cite{paper45} further introduce a semantic branch to enhance FAST-VQA.
With the success of VLMs \cite{paper24,paper111,paper113}, applying them to VQA has become a growing research focus.
For example, CLiF-VQA \cite{paper80} and PTM-VQA \cite{paper108} extract human feeling features from CLIP \cite{paper24} under language supervision, serving as a complement to spatiotemporal features.
In contrast, MaxVQA \cite{paper63} and CLIPVQA \cite{paper124} enhance CLIP’s VQA performance by adapting it to video inputs.


\subsection{Vision-Language Models}
In recent years, advances in VLMs have significantly propelled the development of computer vision.
CLIP \cite{paper24} is a representative model known for its robust representations and strong generalization, learned from comparative training on 400 million image-text pairs.
And then a series of enhanced versions of CLIP are proposed. 
For example, MetaCLIP \cite{paper112} enhances the performance of CLIP by curating and balancing the raw data from the network to improve the quality of the training data.
SigLIP \cite{paper110} replaces the original Softmax with Sigmoid when calculating the loss, leading to improved computational efficiency as well as better performance.
Furthermore, SigLIP2 \cite{paper111}  is trained on a larger-scale dataset and unifies previously disjoint training strategies into a structured, multi-stage pipeline, resulting in notable performance improvements.
Recently, \cite{paper113} train CLIP on a larger image dataset and fine-tune it on a 22M-sized video dataset, significantly enhancing its generalization to video data.
For example, CoOp \cite{paper118}, CLIP-Adapter \cite{paper125}, and Tip-Adapter \cite{paper126} enhance VLMs to enable few-shot image recognition.
Moreover, some works extend VLMs to video tasks.
VideoCLIP \cite{paper127} replaces image-text pairs with video-text pairs for video understanding.
CLIP4Clip \cite{paper128} adapts CLIP for video retrieval by fine-tuning it end-to-end.
ActionCLIP \cite{paper129} and XCLIP \cite{paper130} directly transfer CLIP’s visual representations to video recognition.


\section{Proposed Method}
\begin{figure*}[t]
	\centering
	\includegraphics[scale=0.28]{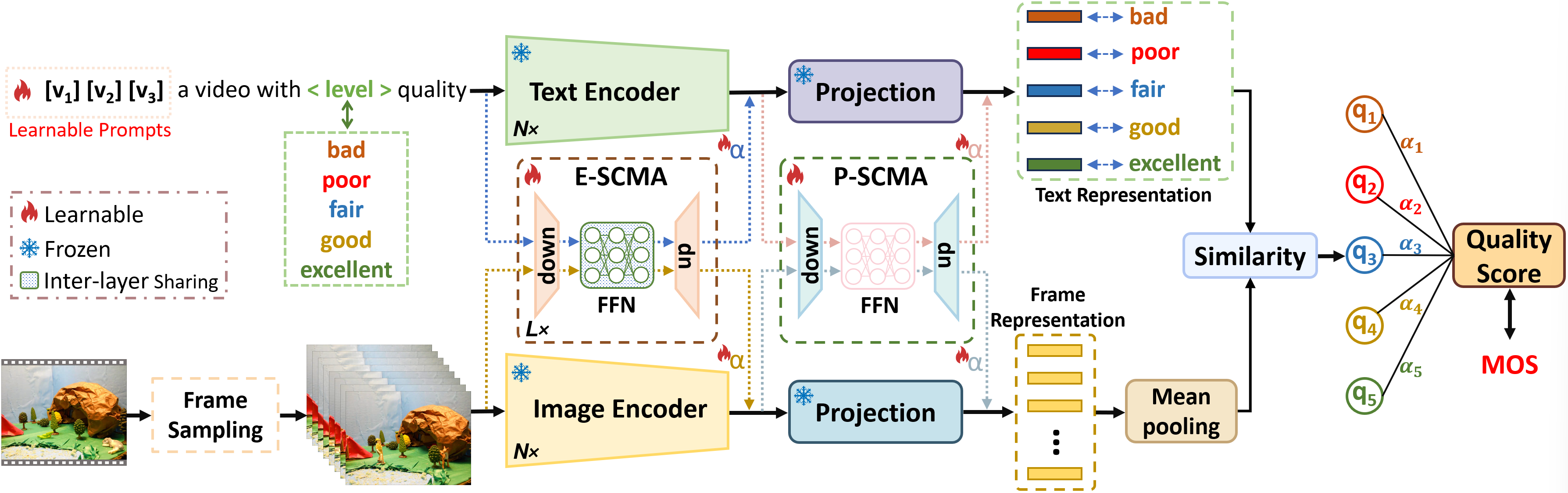}
	\caption{The overall framework of the proposed Q-CLIP.}
	\label{fig1}
\end{figure*}

\begin{figure}[t]
	\centering
	\includegraphics[scale=0.2]{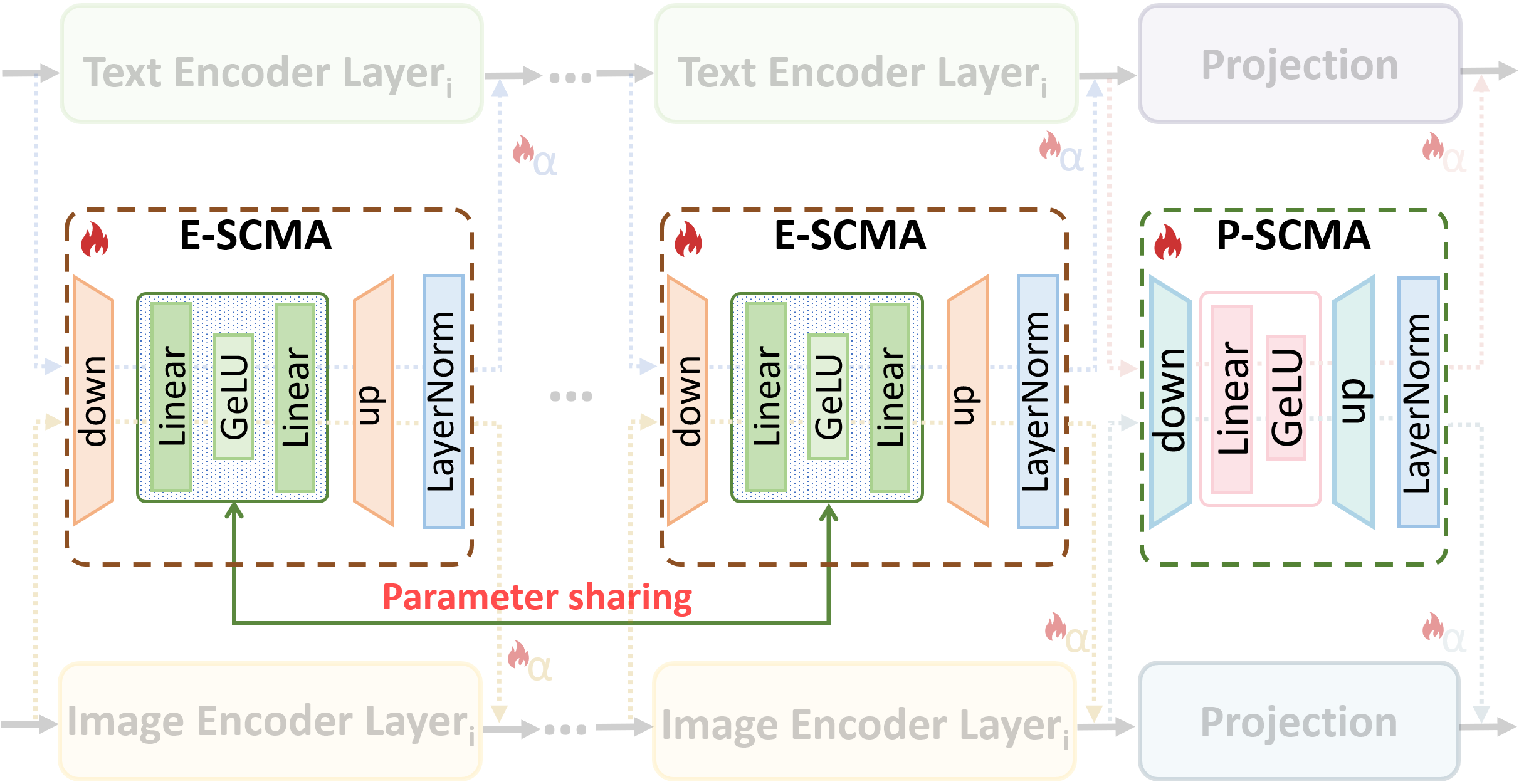}
	\caption{Architecture of the proposed SCMA.}
	\label{fig2}
\end{figure}

\subsection{Overall Architecture}

Our proposed Q-CLIP architecture, illustrated in Fig. \ref{fig1}, is fully built upon the VLMs framework. It enhances the performance of VLMs in VQA by introducing two novel modules: Shared Cross-Modal Adapter (SCMA) and a set of learnable five-level quality prompts.
In addition, we investigate the impact of different frame sampling strategies on VQA performance. 
Beyond conventional random and uniform sampling, we explore a motion-based approach that calculates the difference between each frame and its adjacent frames to measure the intensity of motion. 
Frames are then sampled according to predefined rules based on these motion differences.

\subsubsection{Shared Cross-Modal Adapter}

The core objective of SCMA is to mitigate feature distribution discrepancies between visual and textual branches, facilitating precise cross-modal alignment. 
Fig. \ref{fig2} illustrates the detailed architectural design of SCMA.
Specifically, we design two structures of SCMA, E-SCMA and P-SCMA, to align the features of encoder and projection, respectively.
By reusing the same SCMA architecture for both branches, the model learns a single, generalized strategy to refine and align features, rather than modality-specific heuristics. This consistency ensures that visual and textual features are transformed through comparable operations, reducing the risk of divergent optimization directions that could widen the modality gap.
However, since VLMs contain multiple encoder layers, adding E-SCMA to each of them would lead to a multiplicative increase in the number of trainable parameters, thereby increasing the risk of model overfitting. To address this issue, we incorporate inter-layer parameter sharing in the Feed Forward Network (FFN) of E-SCMA. This design not only effectively reduces training costs but also helps mitigate the risk of overfitting.

Specifically, We denote the inputs to the visual encoders ($E_{k}^{v}$) and text encoders ($E_{k}^{t}$) at layer $k$ as $V_{k}$ and $T_{k}$,  respectively. The frozen encoder extracts features from the inputs to support subsequent inputs:
\begin{equation}
	V_{k}^{'} =E_{k}^{v} (V_{k}), T_{k}^{'} =E_{k}^{t} (T_{k}), 
\end{equation}

For E-SCMA, the processing procedure can be described as follows:
\begin{equation}
	\Delta V_{k} = Up_{k}(FFN(Down_{k}(V_{k}))), 
\end{equation}
\begin{equation}
	\Delta T_{k} = Up_{k}(FFN(Down_{k}(T_{k}))), 
\end{equation}
where $Up_{k}$ and $Down_{k}$ are a linear layer used for dimension expansion and reduction, respectively.
To maintain a consistent feature distribution, LayerNorm is employed for normalization:
\begin{equation}
	\Delta V_{k}^{'}  = LayerNorm(\Delta V_{k}),
\end{equation}
\begin{equation}
	\Delta T_{k}^{'}  = LayerNorm(\Delta T_{k}), 
\end{equation}

Subsequently, the output of E-SCMA is combined with the encoder’s output to obtain the input for the subsequent module:
\begin{equation}
	V_{k+1} = V_{k}^{'}  + \alpha_{k} \Delta V_{k}^{'},
\end{equation}
\begin{equation}
	T_{k+1} = T_{k}^{'}  + \beta_{k} \Delta T_{k}^{'},
\end{equation}
where both $\alpha_{k}$ and $\beta_{k}$ are a learnable parameter.

For P-SCMA, the processing flow is largely similar to that described above. The only difference lies in FFN, which consists of a single linear layer. This is because the projection module is inherently designed to perform simple linear mappings, facilitating similarity computation between the two modalities. To remain consistent with this lightweight mapping structure, we adopt a more streamlined version.

\begin{figure}[t]
	\centering
	\includegraphics[scale=0.25]{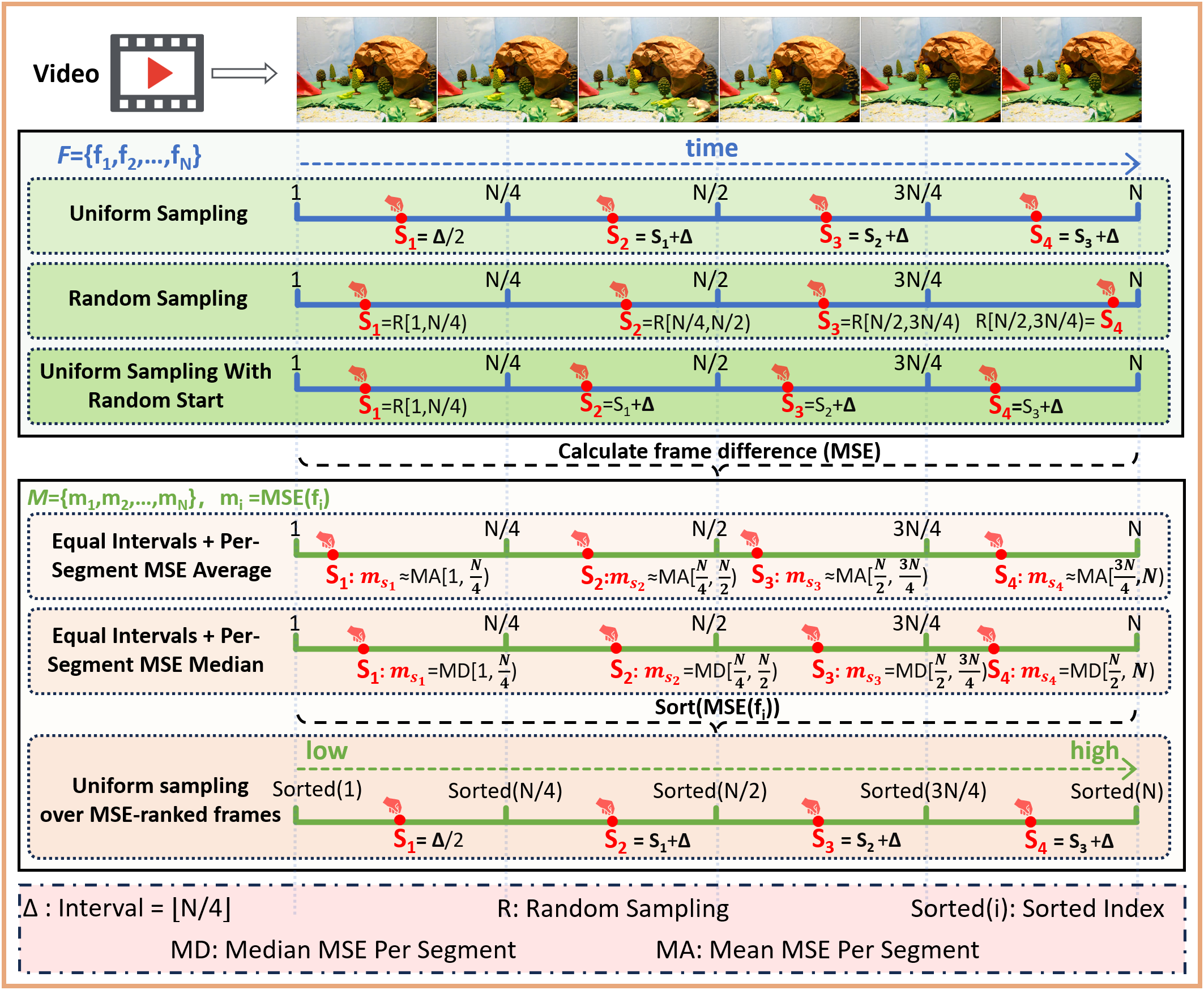}
	\caption{Frame Sampling Diagram.}
	\label{fig1.7}
\end{figure}

\begin{table*}[h]
	
	
	\centering
	\small
	\setlength{\aboverulesep}{0pt}
	\setlength{\belowrulesep}{0pt}
	\centering\setlength{\tabcolsep}{6pt}
	\begin{tabular}{c|c|c|cc|cc|cc|cc}
		\toprule
		\multicolumn{3}{c|}{Testing Type} & \multicolumn{4}{c|}{Intra-dataset Test Datasets} & \multicolumn{4}{c}{Cross-dataset Test Datasets} \\
		\hline
		\multicolumn{3}{c|}{Testing Datasets} & \multicolumn{2}{c|}{\textbf{LSVQ$_{test}$}} & \multicolumn{2}{c|}{\textbf{LSVQ$_{1080p}$}} & \multicolumn{2}{c|}{\textbf{KoNViD-1k}} & \multicolumn{2}{c}{\textbf{LIVE-VQC}} \\
		\hline
		Type & Methods & Source & SROCC & PLCC & SROCC & PLCC & SROCC & PLCC & SROCC & PLCC \\
		\hline
		\multirow{3}{*}{\shortstack{Knowledge-driven}}  & BRISQUE  &\scriptsize \itshape TIP, 2012 &  0.569 & 0.576  & 0.497  & 0.531  & 0.646  & 0.647  & 0.524  & 0.536 \\ 
		& TLVQM  &\scriptsize \itshape TIP, 2019 & 0.772 & 0.774  & 0.589  & 0.616  & 0.732  & 0.724  & 0.670  & 0.691 \\ 
		& VIDEVAL  &\scriptsize \itshape TIP, 2021 &  0.794 & 0.783  & 0.545  & 0.554  & 0.751  & 0.741  & 0.630  & 0.640 \\ 
		\hdashline
		
		\multirow{8}{*}{\shortstack{ Data-driven}} 
		& VSFA & \scriptsize \itshape ACMMM, 2019 & 0.801 & 0.796 & 0.675  & 0.704  & 0.784  & 0.794  & 0.734  & 0.772 \\ 
		& PVQ  &\scriptsize \itshape  CVPR, 2021 &  0.827 & 0.828  & 0.711  & 0.739  & 0.791  & 0.795  & 0.770  & 0.807 \\
		& BVQA  & \scriptsize \itshape TCSVT, 2022 &  0.852 & 0.854  & 0.771  & 0.782  & 0.834  & 0.837  & 0.816  & 0.824 \\
		& FAST-VQA & \scriptsize \itshape ECCV, 2022 &  0.876 & 0.877  & 0.779  & 0.814  & 0.859  & 0.855  & 0.823  & 0.844 \\
		
		& DOVER  & \scriptsize \itshape ICCV, 2023 &  0.881 & 0.879  & 0.782  & 0.827  & 0.871  & 0.872  & 0.812  & 0.841 \\
		& Zoom-VQA & \scriptsize \itshape CVPR,2023 & 0.886 & 0.879 & 0.799  & 0.819 & 0.877 & 0.875 & 0.814 & 0.833 \\
		& MBVQA & \scriptsize \itshape CVPR, 2024 & \underline{0.895}  & 0.895  & 0.809  & 0.844  & 0.878  & 0.884  & 0.806  & 0.844 \\
		
		\hdashline
		\multirow{1}{*}{\shortstack{LLMs}} & Q-Align  & \scriptsize \itshape ICML, 2024 & 0.883  & 0.882  & 0.797  & 0.830  & 0.865  & 0.877  & NA  & NA \\
		\hdashline
		\multirow{10}{*}{\shortstack{VLMs}}
		& PTM-VQA& \scriptsize \itshape CVPR, 2024 &  0.855 & 0.864  & 0.736  & 0.782  & 0.824  & 0.830  & 0.785  & 0.737 \\
		& CLiF-VQA  & \scriptsize \itshape ACMMM, 2024 &  0.886 & 0.887  & 0.790  & 0.832  & 0.877  & 0.874  & \textbf{0.834}  & 0.855 \\
		& CLIPVQA  & \scriptsize \itshape TBC, 2025 &  0.881 & 0.883  & 0.782  & 0.827  & 0.864  & 0.887  & 0.781  & \textbf{0.871} \\

		
		\cline{2-11}
		& \multicolumn{2}{l|}{\textbf{Q-CLIP} \itshape -RandSampl}   & 0.895  & \underline{0.896}  & 0.814  & 0.852 & 0.882  & 0.892  & 0.808  & 0.843 \\
		
		& \multicolumn{2}{l|}{\textbf{Q-CLIP} \itshape -UNISampl}   & \underline{0.897}  & 0.895  & \underline{0.820}  & 0.853 & 0.883  & 0.891  & 0.803  &  0.842 \\
		
		& \multicolumn{2}{l|}{\textbf{Q-CLIP} \itshape -UNIRandStart}   & 0.893  & 0.895  & 0.818  & \underline{0.858}  & 0.883  & 0.890  & 0.804  & 0.844 \\
		
		& \multicolumn{2}{l|}{\textbf{Q-CLIP} \itshape -MSESortedUNI}  & 0.891  & 0.893  & 0.812 & 0.852 & 0.888  & 0.894  &  0.810 & 0.845 \\
		
		& \multicolumn{2}{l|}{\textbf{Q-CLIP} \itshape -SegMSEMean}   & \underline{0.897} & \underline{0.896} & \underline{0.820} & 0.855 & \underline{0.889} & 0.895 & 0.813 & 0.851 \\
		
		& \multicolumn{2}{l|}{\textbf{Q-CLIP} \itshape -SegMSEMedian}  & 0.891 & 0.893 & 0.813 & 0.852 & \underline{0.889} & \underline{0.896}  & 0.813  & 0.852 \\
		
		& \multicolumn{2}{l|}{\textbf{Q-CLIP} \itshape -Mixed}   & \textbf{0.899}  & \textbf{0.900}  & \textbf{0.823}  & \textbf{0.866} & \textbf{0.896}  & \textbf{0.901}  & \underline{0.826}  & \underline{0.867} \\
		\bottomrule
	\end{tabular}
	\caption{Experimental performance of the pre-trained Q-CLIP on the LSVQ dataset on four test sets (LSVQ$_{test}$, LSVQ$_{1080p}$, KoNViD-1k, LIVE-VQC). The best and second best results are bolded and underlined.}
	\label{table1}
\end{table*}

\subsubsection{Learnable Five-level Prompts}
Using antonym-based prompts (e.g., good vs. bad) in VLMs has shown promising results for quality perception. However, since quality assessment is a fine-grained prediction task, such binary prompts are overly coarse and may limit the performance of VLMs in capturing subtle quality differences. Fortunately, recent studies on quality perception using LLMs have demonstrated that converting quality scores into discrete quality levels helps models better capture nuanced hierarchical patterns, leading to improved performance. Inspired by this, we argue that a similar design is also beneficial for VLMs. Therefore, we introduce a prompt scheme with five distinct quality levels:
\begin{equation}
	p = "a\ video\ of" +<level> + "quality"
\end{equation}
Here, $<level>$ represents the five quality levels: {excellent, good, fair, poor, bad}.
However, more specific prompts can often introduce bias into the perception of VLMs. To address this, we introduce learnable prompts to optimize the initial five-level prompt scheme:
\begin{equation}
	\hat{p} = Learnable("X\ X\ X") + p
\end{equation}
We initialize the prompts with three "X" tokens and optimize them during training. All other parts of prompts are keep frozen. The prompts used in this work can be formulated as:
\begin{equation}
	P= \left \{ \hat{p}_{exc},\hat{p}_{good},\hat{p}_{fair},\hat{p}_{poor},\hat{p}_{bad} \right \} 
\end{equation}

\subsubsection{Quality Regression}
The video and text prompts are processed by the model to obtain video features $V$ and text features $T=\left \{ t_{exc},t_{good},t_{fair},t_{poor},t_{bad} \right \} $, respectively. Then, calculate the cosine similarity between the visual content and prompts to predict the score for each dimension:
\begin{equation}
	s_{k} = \frac{t_{k}\cdot V }{\left \| t_{k}  \right \|\left \| V  \right \|}  ,k\in \left \{ exc,good,fair,poor,bad \right \} 
\end{equation}
These similarity scores $ S=\left \{ s_{exc},s_{good},s_{fair},s_{poor},s_{bad} \right \} $ form a quality-level-related distribution, which is further processed to generate the final quality assessment score.
Finally, by applying a weighted sum, the discrete similarity scores are converted into a continuous quality prediction:
\begin{equation}
	 Q_{pred} = {\textstyle \sum_{k=exc}^{bad}} w_{k}\cdot s_{k}
\end{equation}
where $w_{k}$ are learnable weights that calibrate the contribution of each quality level to the final prediction.

\subsubsection{Frame-Difference-Based Sampling}
VQA relies heavily on representative frame samples, as full video sequences are often computationally prohibitive and redundant. 
While random and uniform sampling are widely used as baselines, they overlook the dynamic characteristics of videos that may correlate with quality perception (e.g., motion intensity). 
To address this, we systematically investigate the impact of frame sampling strategies on VQA performance, with a particular focus on frame-difference-based sampling—a strategy rarely explored in prior VQA literature. 
Frame differences, quantified via pixel-wise MSE, reflect motion intensity between consecutive frames:
\begin{equation}
	m_{t} = \frac{1}{2} MSE(v_{i},v_{i+1})+MSE(v_{i},v_{i-1}))
\end{equation}
where $ MSE(a,b)=\frac{1}{H\times W\times 3} {\textstyle \sum_{p}^{}}(a_{p}-b_{p})^{2} $ computes the pixel-wise MSE between frames $a$ and $ b$ with ($p$ indexing individual pixels). The first and last frames are compared only with their single adjacent frames.
Using the frame difference MSE $\left \{ m_{1},m_{2},...m_{N} \right \} $, we design three sampling strategies to select frames, as illustrated in Fig. \ref{fig1.7}, which shows the detailed sampling process.

\section{Experiments}

\begin{table*}[t]
	\small
	\setlength{\aboverulesep}{0pt}
	\setlength{\belowrulesep}{0pt}
	\centering\setlength{\tabcolsep}{5pt}
	
	\begin{tabular}{c|c|c|cc|cc|cc|cc|cc}
		\toprule
		\multicolumn{3}{c|}{Finetune Datasets} & \multicolumn{2}{c|}{\textbf{LIVE-VQC}} & \multicolumn{2}{c|}{\textbf{KoNViD-1k}} & \multicolumn{2}{c|}{\textbf{YouTube-UGC}} & \multicolumn{2}{c|}{\textbf{CVD2014}} &\multicolumn{2}{c}{\textbf{LIVE-Qualcomm}} \\
		\hline
		Type & Methods & Source & SRCC & PLCC & SRCC & PLCC & SRCC & PLCC & SRCC & PLCC & SRCC & PLCC   \\
		\hline
		\multirow{3}{*}{\shortstack{Knowledge- \\ driven}}
		& TLVQM & \scriptsize \itshape TIP, 2019 & 0.799 & 0.803  & 0.773  & 0.768  & 0.669  & 0.659 & 0.830 & 0.850 & 0.770 & 0.810 \\ 
		& VIDEVAL & \scriptsize \itshape TIP, 2021 &  0.752 & 0.751  & 0.783  & 0.780  & 0.779  & 0.773 & NA & NA & NA & NA  \\ 
		& RAPIQUE  & \scriptsize \itshape OJSP, 2021 &  0.755 & 0.786  & 0.803  & 0.817  & 0.759  & 0.768 & NA & NA & NA & NA   \\
		\hdashline
		\multirow{9}{*}{\shortstack{Data-driven}} 
		& VSFA & \scriptsize \itshape ACMMM, 2019 & 0.773 & 0.795 & 0.773  & 0.775  & 0.724  & 0.743 & 0.870 & 0.868 & 0.737 & 0.732\\ 
		& GST-VQA & \scriptsize \itshape TCSVT, 2021 & NA & NA & 0.814 & 0.825 & NA & NA & 0.831 & 0.844 & 0.801 & 0.825 \\ 
		& PVQ  &  \scriptsize \itshape CVPR, 2021 &  0.827 & 0.837  & 0.791  & 0.786  & NA  & NA  & NA & NA & NA & NA\\ 
		& BVQA  & \scriptsize \itshape TCSVT, 2022 &  0.841 & 0.839  & 0.835  & 0.834  & 0.825  & 0.818 & 0.863 & 0.883 & 0.833 & 0.837  \\
		& CoINVQ  & \scriptsize \itshape TCSVT, 2021 & NA & NA & 0.767 & 0.764 & 0.816 & 0.802 & NA & NA & NA & NA  \\
		& FAST-VQA  & \scriptsize \itshape ECCV, 2022 &  0.845 & 0.852  & 0.890  & 0.889  &  0.857  & 0.853  & \underline{0.891} & \underline{0.903} & 0.819 & 0.851\\
		& DOVER  & \scriptsize \itshape ICCV, 2023 &  0.812 & 0.852  & 0.897  &0.899  & 0.877  & 0.873 & 0.858 & 0.881 & 0.736 & 0.789\\
		& MBVQA  & \scriptsize \itshape CVPR, 2024 &  0.860 & 0.880 & 0.901  & 0.905  & 0.876  & 0.877 & 0.883 & 0.901 & 0.832 & 0.842\\
		\hdashline
		\multirow{5}{*}{\shortstack{VLMs}} 
		& MaxVQA  & \scriptsize \itshape ACMMM, 2023 & 0.854 & 0.873 & 0.894 & 0.895 & 0.894 & \underline{0.890} & NA & NA & NA & NA \\
		& PTM-VQA& \scriptsize \itshape CVPR, 2024 &  0.811 & 0.820  & 0.857  & 0.872  & 0.858  & 0.857 & NA & NA & NA & NA  \\
		& CLiF-VQA  & \scriptsize \itshape ACMMM, 2024 &  0.866 & 0.878  & 0.903  & 0.903  & \underline{0.888}  & \underline{0.890} & 0.881 & 0.891 & 0.832 & 0.850 \\
		& CLIPVQA  & \scriptsize \itshape TBC, 2025 &  \underline{0.870} & \underline{0.892}  & \underline{0.907}  & \underline{0.912}  & 0.881  & 0.883 & 0.883  & 0.888 & \underline{0.833}  & \underline{0.872} \\

		\cline{2-13}
		& \textbf{Q-CLIP} &  \textbf{\itshape Ours} & \textbf{0.881} & \textbf{0.901} & \textbf{0.915} & \textbf{0.920} & \textbf{0.911} & \textbf{0.911} & \textbf{0.897} & \textbf{0.907}  & \textbf{0.846} & \textbf{0.884} \\ 
		
		
		\bottomrule
	\end{tabular}
	\caption{The finetune results on LIVE-VQC, KoNViD-1k, YouTube-UGC, CVD2014 and LIVE-Qualcomm.
		The best and second best results are bolded and underlined.
	}
	\label{table2}
\end{table*}

\subsection{Experimental Setups}
\subsubsection{Datasets}
We verify our model on six datasets: LSVQ \cite{paper16}, KoNViD-1k (1200) \cite{paper1}, LIVE-VQC (585) \cite{paper3}, YouTube-UGC (1067) \cite{paper6}, CVD2014 (234) \cite{paper2}, LIVE-Qualcomm (208) \cite{paper107}. 
We pre-train Q-CLIP on LSVQ (28056), with intra-dataset testing on LSVQ$_{test}$ (7400) and LSVQ$_{1080p}$ (3600), and cross-dataset testing on KoNViD-1k and LIVE-VQC.
Further, we fine-tune the model on KoNViD-1k, LIVE-VQC, YouTube-UGC, CVD2014 and LIVE-Qualcomm.
Following standard practice, we split each dataset into 80\% training and 20\% testing, and report the average performance over ten random splits.

\subsubsection{Evaluation Criteria}
The Spearman Rank Order Correlation Coefficient (SROCC), the Kendall Rank Order Correlation Coefficient (KROCC), the Pearson Linear Correlation Coefficient (PLCC), and the Root Mean Square Error (RMSE) are used as evaluation Metrics. Higher SROCC, PLCC, KROCC and lower RMSE scores represent models with better performance.

\subsubsection{Implementation Details}
we employ PyTorch framework and an NVIDIA GeForce RTX 4090 card to train the model in all experimental implementations.
As most current VLMs are trained on static images, they are not well-equipped to model the temporal dynamics in videos. To address this, we adopt a CLIP variant (PE$_{core}$-L) \cite{paper113} that has been pre-tuned on video data as our backbone. 
We sample 8 frames per video as input.
We set the initial learning rate to 0.001, the optimizer to AdamW, and use a cosine annealing strategy to dynamically adjust the learning rate. 
And training is conducted for 8 epochs using a batch size of 12.

\subsection{Pre-training Results on LSVQ}
We pre-train the proposed Q-CLIP on LSVQ and conduct intra-dataset testing on LSVQ$_{test}$ and LSVQ$_{1080p}$. Additionally, cross-dataset testing was performed on KoNViD-1k and LIVE-VQC. 
Furthermore, we examine the effectiveness of various frame sampling methods.
The results are shown in Tab. \ref{table1}.
Frame-difference-based samplings achieve comparable results to traditional methods, such as random and uniform sampling, in intra-dataset testing. In cross-dataset testing, frame-difference-based samplings demonstrate superior performance compared to traditional approaches, indicating better generalization capability.
This suggests that frame-difference-based samplings can more effectively select frames that are representative of video quality, particularly in cross-dataset scenarios. In contrast, traditional sampling methods do not consider any intrinsic characteristics of the video frames, which may result in redundant or less informative samples, thereby limiting their representativeness and generalizability.
The results indicate that regardless of the sampling strategy employed, Q-CLIP consistently achieves state-of-the-art performance. Moreover, integrating multiple sampling methods during training further enhances the model’s overall performance.


Compared with knowledge-driven and data-driven methods, Q-CLIP achieves a significant improvement over all datasets.
Furthermore, it also demonstrates distinct advantages over Q-Align, which is based on LLMs.
Compared with VLM-based methods, although Q-CLIP performs slightly lower than CLIF-VQA and CLIPVQA on the LIVE-VQC dataset, it significantly outperforms them on the other three datasets.
Specifically, Q-CLIP improves over CLiF-VQA and CLIPVQA by up to 3.7\% in SROCC and 2.9\% in PLCC.

\begin{figure}[htbp]
	\centering
	\includegraphics[scale=0.26]{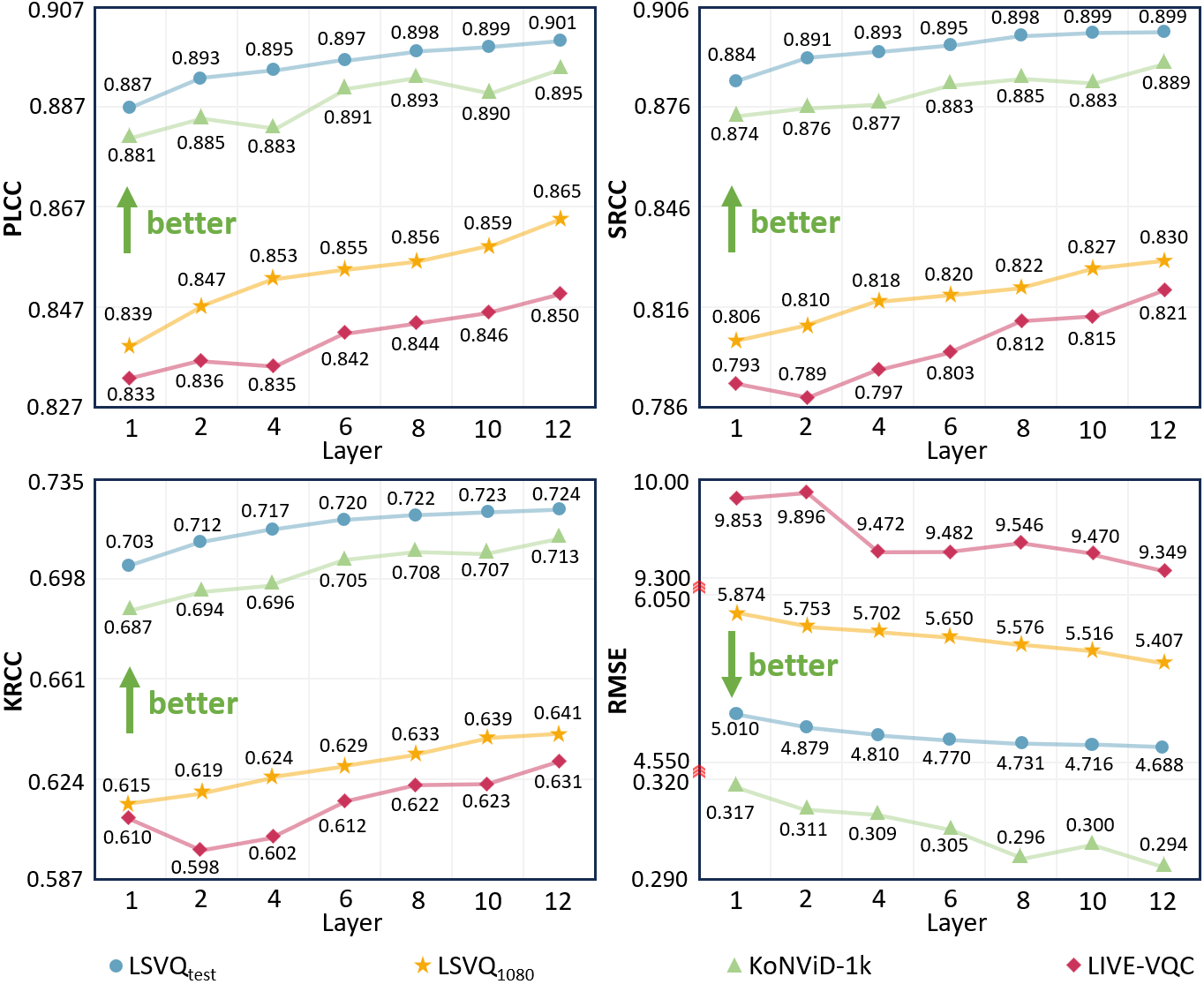}
	\caption{Ablation on the number of E-SCMA Layers.}
	\label{fig3}
\end{figure}

\subsection{Fine-tuning Results on Small Datasets}
After pre-training on LSVQ, we fine-tune Q-CLIP on five small datasets (LIVE-VQC, KoNViD-1k, YouTube-UGC, CVD2014, LIVE-Qualcomm), as shown in Tab. \ref{table2}. 
Specifically, we use uniform sampling as the sampling strategy during fine-tuning.
As can be seen, Q-CLIP achieves unprecedented performance on all five datasets.
Compared to the current best performance, Q-CLIP improves the average performance on SROCC and PLCC by 1.39\% and 1.22\%, respectively. 
Furthermore, Q-CLIP outperforms the state-of-the-art VLMs-based method CLIPVQA by 1.74\% and 1.72\%  in SROCC and PLCC, respectively.
The results further illustrate the validity of our proposed Q-CLIP.

\subsection{Ablation Studies}
We conduct experimental analysis to evaluate the effectiveness of each component. 
Specifically, ablation experiments are by default based on a uniform sampling strategy.

\begin{table}[h]
	\centering
	\setlength{\tabcolsep}{5.5pt}
	\setlength{\aboverulesep}{0pt}
	\setlength{\belowrulesep}{0pt}
	\small
	\begin{tabular}{cccc|cc}
		\toprule
		Visual & Text & Sharing & Layer sharing & SROCC & PLCC \\
		\hline
		\checkmark &   &   &  & 0.866 &  0.864 \\
		& \checkmark  &   &   & 0.837 & 0.838 \\
		\checkmark & \checkmark  &   &   & 0.875 & 0.878 \\
		\checkmark & \checkmark  & \checkmark &   & 0.885 & 0.886 \\
		\checkmark & \checkmark  & \checkmark & \checkmark  & \textbf{0.895} & \textbf{0.897} \\
		\bottomrule
	\end{tabular}
	\caption{Ablation on the structure of SCMA.}
	\label{table3}
\end{table}

\begin{table}[htbp]
	\centering
	\setlength{\tabcolsep}{2.5pt}
	\small
	\setlength{\aboverulesep}{0pt}
	\setlength{\belowrulesep}{0pt}
	\begin{tabular}{l|cc|cc|cc}
		\toprule
		Datasets &\multicolumn{2}{c|}{\textbf{LSVQ$_{test}$}} & \multicolumn{2}{c|}{\textbf{KoNViD-1k}} & \multicolumn{2}{c}{\textbf{LIVE-VQC}}   \\
		\hline
		Prompts & SROCC & PLCC & SROCC & PLCC & SROCC & PLCC  \\
		\hline
		\itshape Antonym & 0.883 & 0.881 & 0.866 & 0.867 & 0.791 & 0.820 \\
		\itshape Antonym{*} & 0.885 & 0.883 & 0.871 & 0.879 & 0.789 & 0.826 \\
		\itshape Five levels & 0.891 & 0.891 & 0.874 & 0.885 & 0.798 & 0.837 \\
		\itshape Five levels{*} & \textbf{0.895} & \textbf{0.897} & \textbf{0.883} & \textbf{0.891} & \textbf{0.803} & \textbf{0.842} \\
		%
		\bottomrule
	\end{tabular}
	\caption{Ablation on prompts. {*} represents learnable.}
	\label{table4}
\end{table}

\subsubsection{Ablation on SCMA}
We validate the effectiveness of SCMA on LSVQ, as shown in Tab. \ref{table3}.
Applying SCMA to either the visual or textual branch individually results in limited performance. When SCMA is applied to both branches without parameter sharing, the performance improves notably. Sharing SCMA across the two branches leads to further gains, and the best results are achieved when inter-layer sharing is additionally introduced. These results validate the effectiveness of our proposed SCMA architecture, which jointly leverages branch-wise and inter-layer sharing.

Furthermore, since VLMs typically consist of multiple layers, we further investigated the impact of the number of inserted E-SCMA layers on model performance. As shown in Fig. \ref{fig3}, we train the model on LSVQ$_{train}$ and evaluate it on LSVQ$_{test}$, LSVQ$_{1080p}$, KoNViD-1k, and LIVE-VQC. The results demonstrate that as the number of E-SCMA layers increases, the model performance consistently improves. Notably, all comparative experiments in this paper are based on a 6-layer E-SCMA configuration, suggesting that further performance gains can be achieved by increasing the number of E-SCMA layers.

\subsubsection{Ablation on Prompts}
Most existing VLMs-based quality assessment methods \cite{paper63,paper25} utilize antonym-based prompts. 
To validate the effectiveness of our proposed prompts, we compare it against the antonym-based prompts.
As shown in Tab. \ref{table4}. 
Compared to antonym-based prompts, our five-level prompt design offers a clear advantage. Furthermore, performance is further improved by introducing learnable parameters.

\begin{figure}[htbp]
	\centering
	\includegraphics[scale=0.25]{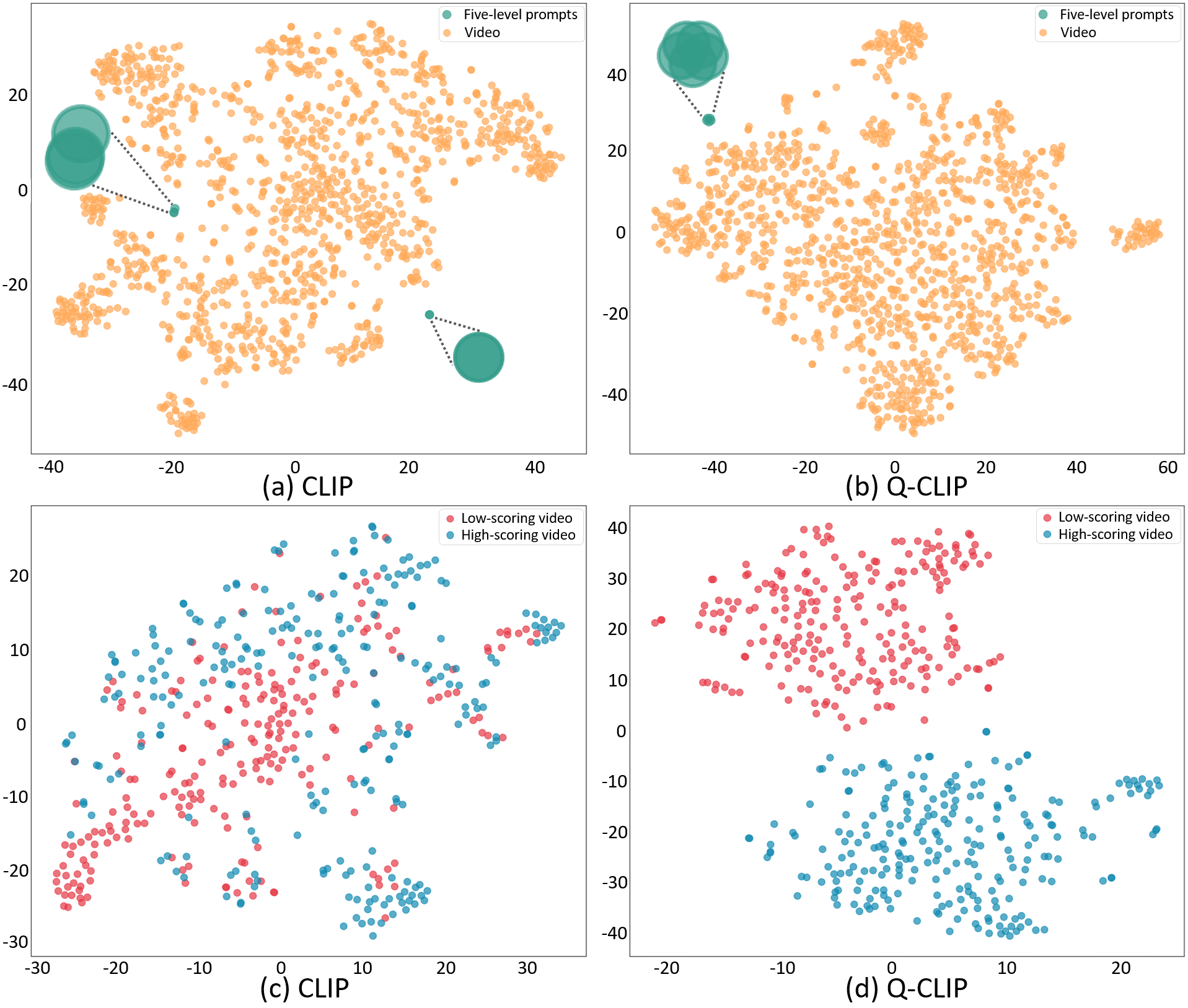}
	\caption{t-SNE visualizations on KoNViD-1k dataset.}
	\label{fig6}
\end{figure}

%

\subsection{Efficiency}
Compared to models specifically designed for VQA, VLMs typically have a larger number of parameters, making model efficiency a critical consideration.
To this end, we conduct extensive experiments to verify the efficiency of Q-CLIP.
We first compare the total and fine-tuned parameters with several state-of-the-art VQA methods, as shown in Fig. \ref{fig5}.
Q-CLIP requires only a minimal number of fine-tuned parameters \textbf{(0.14M)} to achieve top performance.
For example, compared with MBVQA, CLIP-VQA, and Q-Align, Q-CLIP reduces the fine-tuned parameters by \textbf{666$\times$}, \textbf{3964$\times$}, and \textbf{58556$\times$} respectively.
Moreover, even when compared to the most efficient method, FAST-VQA, Q-CLIP still achieves a \textbf{197$\times$} reduction in fine-tuned parameters.
Additionally, Q-CLIP’s total parameters remain reasonable.
Compared to CLIPVQA, which is also based on VLMs, Q-CLIP has a comparable total parameter size. 
In contrast, Q-CLIP reduces the total parameters by \textbf{13$\times$} compared to Q-Align, which is based on LLMs.

%
\subsection{Visualization}
To further evaluate Q-CLIP’s quality perception capability, we visualize feature distributions using t-SNE (Fig. \ref{fig6}). 
From Fig. 6a and Fig. 6b, the original CLIP shows clear modality confusion, as visual and textual features overlap significantly. Additionally, the five-level prompt features are scattered and poorly separated, with some quality levels fully overlapping. This indicates that CLIP fails to effectively encode quality-relevant information.
In contrast, Q-CLIP exhibits a well-structured feature space, where visual and textual modalities are distinctly separated, and quality levels form compact, aligned clusters. Such separability is key to robust cross-modal understanding in high-performing VLMs \cite{paper115,paper116,paper117}.
Further analysis of video features by quality (Fig. 6c and Fig. 6d) shows that Q-CLIP clearly distinguishes high- and low-scoring videos, unlike CLIP, which presents significant overlap. These results confirm that Q-CLIP enhances the feature space through effective modality separation and structured quality encoding, enabling accurate quality perception for VQA.

\section{Conclusion}
In this paper, we introduce Q-CLIP, the first VQA framework built entirely upon VLMs.
A Shared Cross-Modal Adapter (SCMA) is employed to optimize feature representations in both the visual and textual branches. Thanks to its minimal number of trainable parameters, SCMA significantly reduces the training cost. Additionally, a learnable five-level prompt mechanism is introduced to help the model perceive fine-grained quality variations.
Furthermore, we investigate the impact of different frame sampling strategies on VQA.
Experimental results demonstrate that Q-CLIP outperforms existing methods on multiple VQA datasets.

\bibliography{main}

\begin{thebibliography}{65}
\providecommand{\natexlab}[1]{#1}

\bibitem[{Bolya et~al.(2025)Bolya, Huang, Sun, Cho, Madotto, Wei, Ma, Zhi,
  Rajasegaran, Rasheed et~al.}]{paper113}
Bolya, D.; Huang, P.-Y.; Sun, P.; Cho, J.~H.; Madotto, A.; Wei, C.; Ma, T.;
  Zhi, J.; Rajasegaran, J.; Rasheed, H.; et~al. 2025.
\newblock Perception encoder: The best visual embeddings are not at the output
  of the network.
\newblock \emph{arXiv preprint arXiv:2504.13181}.

\bibitem[{Chen et~al.(2021)Chen, Zhu, Li, Lu, Fan, and Wang}]{paper41}
Chen, B.; Zhu, L.; Li, G.; Lu, F.; Fan, H.; and Wang, S. 2021.
\newblock Learning generalized spatial-temporal deep feature representation for
  no-reference video quality assessment.
\newblock \emph{IEEE TCSVT}, 32(4): 1903--1916.

\bibitem[{Cho et~al.(2014)Cho, Van~Merri{\"e}nboer, Gulcehre, Bahdanau,
  Bougares, Schwenk, and Bengio}]{paper58}
Cho, K.; Van~Merri{\"e}nboer, B.; Gulcehre, C.; Bahdanau, D.; Bougares, F.;
  Schwenk, H.; and Bengio, Y. 2014.
\newblock Learning phrase representations using RNN encoder-decoder for
  statistical machine translation.
\newblock \emph{arXiv preprint arXiv:1406.1078}.

\bibitem[{Deng et~al.(2009)Deng, Dong, Socher, Li, Li, and Fei-Fei}]{paper57}
Deng, J.; Dong, W.; Socher, R.; Li, L.-J.; Li, K.; and Fei-Fei, L. 2009.
\newblock Imagenet: A large-scale hierarchical image database.
\newblock In \emph{CVPR}, 248--255.

\bibitem[{Gao et~al.(2024)Gao, Geng, Zhang, Ma, Fang, Zhang, Li, and
  Qiao}]{paper125}
Gao, P.; Geng, S.; Zhang, R.; Ma, T.; Fang, R.; Zhang, Y.; Li, H.; and Qiao, Y.
  2024.
\newblock Clip-adapter: Better vision-language models with feature adapters.
\newblock \emph{IJCV}, 132(2): 581--595.

\bibitem[{Ghadiyaram et~al.(2017)Ghadiyaram, Pan, Bovik, Moorthy, Panda, and
  Yang}]{paper107}
Ghadiyaram, D.; Pan, J.; Bovik, A.~C.; Moorthy, A.~K.; Panda, P.; and Yang,
  K.-C. 2017.
\newblock In-capture mobile video distortions: A study of subjective behavior
  and objective algorithms.
\newblock \emph{IEEE TCSVT}, 28(9): 2061--2077.

\bibitem[{Hara, Kataoka, and Satoh(2017)}]{paper52}
Hara, K.; Kataoka, H.; and Satoh, Y. 2017.
\newblock Learning spatio-temporal features with 3d residual networks for
  action recognition.
\newblock In \emph{ICCV Workshops}, 3154--3160.

\bibitem[{Hara, Kataoka, and Satoh(2018)}]{paper51}
Hara, K.; Kataoka, H.; and Satoh, Y. 2018.
\newblock Can spatiotemporal 3d cnns retrace the history of 2d cnns and
  imagenet?
\newblock In \emph{CVPR}, 6546--6555.

\bibitem[{He et~al.(2016)He, Zhang, Ren, and Sun}]{paper48}
He, K.; Zhang, X.; Ren, S.; and Sun, J. 2016.
\newblock Deep residual learning for image recognition.
\newblock In \emph{CVPR}, 770--778.

\bibitem[{Hosu et~al.(2017)Hosu, Hahn, Jenadeleh, Lin, Men, Szir{\'a}nyi, Li,
  and Saupe}]{paper1}
Hosu, V.; Hahn, F.; Jenadeleh, M.; Lin, H.; Men, H.; Szir{\'a}nyi, T.; Li, S.;
  and Saupe, D. 2017.
\newblock The Konstanz natural video database (KoNViD-1k).
\newblock In \emph{QoMEX}, 1--6. IEEE.

\bibitem[{Jia et~al.(2021)Jia, Yang, Xia, Chen, Parekh, Pham, Le, Sung, Li, and
  Duerig}]{paper109}
Jia, C.; Yang, Y.; Xia, Y.; Chen, Y.-T.; Parekh, Z.; Pham, H.; Le, Q.; Sung,
  Y.-H.; Li, Z.; and Duerig, T. 2021.
\newblock Scaling up visual and vision-language representation learning with
  noisy text supervision.
\newblock In \emph{ICML}, 4904--4916. PMLR.

\bibitem[{Ju et~al.(2022)Ju, Han, Zheng, Zhang, and Xie}]{paper121}
Ju, C.; Han, T.; Zheng, K.; Zhang, Y.; and Xie, W. 2022.
\newblock Prompting visual-language models for efficient video understanding.
\newblock In \emph{ECCV}, 105--124.

\bibitem[{Kay et~al.(2017)Kay, Carreira, Simonyan, Zhang, Hillier,
  Vijayanarasimhan, Viola, Green, Back, Natsev et~al.}]{paper65}
Kay, W.; Carreira, J.; Simonyan, K.; Zhang, B.; Hillier, C.; Vijayanarasimhan,
  S.; Viola, F.; Green, T.; Back, T.; Natsev, P.; et~al. 2017.
\newblock The kinetics human action video dataset.
\newblock \emph{arXiv preprint arXiv:1705.06950}.

\bibitem[{Korhonen(2019)}]{paper8}
Korhonen, J. 2019.
\newblock Two-level approach for no-reference consumer video quality
  assessment.
\newblock \emph{IEEE TIP}, 28(12): 5923--5938.

\bibitem[{Li et~al.(2022)Li, Zhang, Tian, Zhai, and Wang}]{paper17}
Li, B.; Zhang, W.; Tian, M.; Zhai, G.; and Wang, X. 2022.
\newblock Blindly assess quality of in-the-wild videos via quality-aware
  pre-training and motion perception.
\newblock \emph{IEEE TCSVT}, 32(9): 5944--5958.

\bibitem[{Li, Jiang, and Jiang(2019)}]{paper15}
Li, D.; Jiang, T.; and Jiang, M. 2019.
\newblock Quality assessment of in-the-wild videos.
\newblock In \emph{ACM MM}, 2351--2359.

\bibitem[{Liang et~al.(2022)Liang, Zhang, Kwon, Yeung, and Zou}]{paper115}
Liang, V.~W.; Zhang, Y.; Kwon, Y.; Yeung, S.; and Zou, J.~Y. 2022.
\newblock Mind the gap: Understanding the modality gap in multi-modal
  contrastive representation learning.
\newblock \emph{NIPS}, 35: 17612--17625.

\bibitem[{Liu et~al.(2023)Liu, Wu, Yuan, Sun, Tang, Zheng, Wen, and
  Li}]{paper132}
Liu, H.; Wu, M.; Yuan, K.; Sun, M.; Tang, Y.; Zheng, C.; Wen, X.; and Li, X.
  2023.
\newblock Ada-dqa: Adaptive diverse quality-aware feature acquisition for video
  quality assessment.
\newblock In \emph{ACM MM}, 6695--6704.

\bibitem[{Liu et~al.(2022)Liu, Ning, Cao, Wei, Zhang, Lin, and Hu}]{paper53}
Liu, Z.; Ning, J.; Cao, Y.; Wei, Y.; Zhang, Z.; Lin, S.; and Hu, H. 2022.
\newblock Video swin transformer.
\newblock In \emph{CVPR}, 3202--3211.

\bibitem[{Lu et~al.(2022)Lu, Liu, Zhang, Liu, and Tian}]{paper120}
Lu, Y.; Liu, J.; Zhang, Y.; Liu, Y.; and Tian, X. 2022.
\newblock Prompt distribution learning.
\newblock In \emph{CVPR}, 5206--5215.

\bibitem[{Luo et~al.(2022)Luo, Ji, Zhong, Chen, Lei, Duan, and Li}]{paper128}
Luo, H.; Ji, L.; Zhong, M.; Chen, Y.; Lei, W.; Duan, N.; and Li, T. 2022.
\newblock Clip4clip: An empirical study of clip for end to end video clip
  retrieval and captioning.
\newblock \emph{Neurocomputing}, 508: 293--304.

\bibitem[{Mi et~al.(2024{\natexlab{a}})Mi, Li, Shu, and Liu}]{paper81}
Mi, Y.; Li, Y.; Shu, Y.; and Liu, S. 2024{\natexlab{a}}.
\newblock {ZE-FESG}: A Zero-Shot Feature Extraction Method Based on Semantic
  Guidance for No-Reference Video Quality Assessment.
\newblock In \emph{ICASSP}, 3640--3644.

\bibitem[{Mi et~al.(2024{\natexlab{b}})Mi, Shu, Li, Hui, Zhou, and
  Liu}]{paper80}
Mi, Y.; Shu, Y.; Li, Y.; Hui, C.; Zhou, P.; and Liu, S. 2024{\natexlab{b}}.
\newblock {CLiF-VQA}: Enhancing Video Quality Assessment by Incorporating
  High-Level Semantic Information related to Human Feelings.
\newblock In \emph{ACM MM}, 9989–9998.

\bibitem[{Mittal, Moorthy, and Bovik(2012)}]{paper30}
Mittal, A.; Moorthy, A.~K.; and Bovik, A.~C. 2012.
\newblock No-reference image quality assessment in the spatial domain.
\newblock \emph{IEEE TIP}, 21(12): 4695--4708.

\bibitem[{Mittal, Saad, and Bovik(2015)}]{paper10}
Mittal, A.; Saad, M.~A.; and Bovik, A.~C. 2015.
\newblock A completely blind video integrity oracle.
\newblock \emph{IEEE TIP}, 25(1): 289--300.

\bibitem[{Ni et~al.(2022)Ni, Peng, Chen, Zhang, Meng, Fu, Xiang, and
  Ling}]{paper130}
Ni, B.; Peng, H.; Chen, M.; Zhang, S.; Meng, G.; Fu, J.; Xiang, S.; and Ling,
  H. 2022.
\newblock Expanding language-image pretrained models for general video
  recognition.
\newblock In \emph{ECCV}, 1--18.

\bibitem[{Nuutinen et~al.(2016)Nuutinen, Virtanen, Vaahteranoksa, Vuori,
  Oittinen, and H{\"a}kkinen}]{paper2}
Nuutinen, M.; Virtanen, T.; Vaahteranoksa, M.; Vuori, T.; Oittinen, P.; and
  H{\"a}kkinen, J. 2016.
\newblock {CVD2014—A} database for evaluating no-reference video quality
  assessment algorithms.
\newblock \emph{IEEE TIP}, 25(7): 3073--3086.

\bibitem[{Qian, Xu, and Hu(2023)}]{paper116}
Qian, Q.; Xu, Y.; and Hu, J. 2023.
\newblock Intra-modal proxy learning for zero-shot visual categorization with
  clip.
\newblock \emph{NIPS}, 36: 25461--25474.

\bibitem[{Radford et~al.(2021)Radford, Kim, Hallacy, Ramesh, Goh, Agarwal,
  Sastry, Askell, Mishkin, Clark et~al.}]{paper24}
Radford, A.; Kim, J.~W.; Hallacy, C.; Ramesh, A.; Goh, G.; Agarwal, S.; Sastry,
  G.; Askell, A.; Mishkin, P.; Clark, J.; et~al. 2021.
\newblock Learning transferable visual models from natural language
  supervision.
\newblock In \emph{ICML}, 8748--8763. PMLR.

\bibitem[{Saad, Bovik, and Charrier(2014)}]{paper11}
Saad, M.~A.; Bovik, A.~C.; and Charrier, C. 2014.
\newblock Blind prediction of natural video quality.
\newblock \emph{IEEE TIP}, 23(3): 1352--1365.

\bibitem[{Simonyan and Zisserman(2014)}]{paper47}
Simonyan, K.; and Zisserman, A. 2014.
\newblock Very deep convolutional networks for large-scale image recognition.
\newblock \emph{arXiv preprint arXiv:1409.1556}.

\bibitem[{Sinno and Bovik(2018)}]{paper3}
Sinno, Z.; and Bovik, A.~C. 2018.
\newblock Large-scale study of perceptual video quality.
\newblock \emph{IEEE TIP}, 28(2): 612--627.

\bibitem[{Sun et~al.(2022)Sun, Min, Lu, and Zhai}]{paper18}
Sun, W.; Min, X.; Lu, W.; and Zhai, G. 2022.
\newblock A deep learning based no-reference quality assessment model for ugc
  videos.
\newblock In \emph{ACM MM}, 856--865.

\bibitem[{Tran et~al.(2015)Tran, Bourdev, Fergus, Torresani, and
  Paluri}]{paper50}
Tran, D.; Bourdev, L.; Fergus, R.; Torresani, L.; and Paluri, M. 2015.
\newblock Learning spatiotemporal features with 3d convolutional networks.
\newblock In \emph{ICCV}, 4489--4497.

\bibitem[{Tschannen et~al.(2025)Tschannen, Gritsenko, Wang, Naeem,
  Alabdulmohsin, Parthasarathy, Evans, Beyer, Xia, Mustafa et~al.}]{paper111}
Tschannen, M.; Gritsenko, A.; Wang, X.; Naeem, M.~F.; Alabdulmohsin, I.;
  Parthasarathy, N.; Evans, T.; Beyer, L.; Xia, Y.; Mustafa, B.; et~al. 2025.
\newblock Siglip 2: Multilingual vision-language encoders with improved
  semantic understanding, localization, and dense features.
\newblock \emph{arXiv preprint arXiv:2502.14786}.

\bibitem[{Tu et~al.(2021{\natexlab{a}})Tu, Wang, Birkbeck, Adsumilli, and
  Bovik}]{paper7}
Tu, Z.; Wang, Y.; Birkbeck, N.; Adsumilli, B.; and Bovik, A.~C.
  2021{\natexlab{a}}.
\newblock {UGC-VQA:} Benchmarking blind video quality assessment for user
  generated content.
\newblock \emph{IEEE TIP}, 30: 4449--4464.

\bibitem[{Tu et~al.(2021{\natexlab{b}})Tu, Yu, Wang, Birkbeck, Adsumilli, and
  Bovik}]{paper66}
Tu, Z.; Yu, X.; Wang, Y.; Birkbeck, N.; Adsumilli, B.; and Bovik, A.~C.
  2021{\natexlab{b}}.
\newblock RAPIQUE: Rapid and accurate video quality prediction of user
  generated content.
\newblock \emph{IEEE OJSP}, 2: 425--440.

\bibitem[{Wang, Chan, and Loy(2023)}]{paper25}
Wang, J.; Chan, K.~C.; and Loy, C.~C. 2023.
\newblock Exploring clip for assessing the look and feel of images.
\newblock In \emph{AAAI}, volume~37, 2555--2563.

\bibitem[{Wang et~al.(2023)Wang, Xing, Mei, Liu, and Jiang}]{paper129}
Wang, M.; Xing, J.; Mei, J.; Liu, Y.; and Jiang, Y. 2023.
\newblock Actionclip: Adapting language-image pretrained models for video
  action recognition.
\newblock \emph{IEEE TNNLS}.

\bibitem[{Wang, Inguva, and Adsumilli(2019)}]{paper6}
Wang, Y.; Inguva, S.; and Adsumilli, B. 2019.
\newblock YouTube UGC dataset for video compression research.
\newblock In \emph{MMSP}, 1--5.

\bibitem[{Wang et~al.(2021)Wang, Ke, Talebi, Yim, Birkbeck, Adsumilli,
  Milanfar, and Yang}]{paper22}
Wang, Y.; Ke, J.; Talebi, H.; Yim, J.~G.; Birkbeck, N.; Adsumilli, B.;
  Milanfar, P.; and Yang, F. 2021.
\newblock Rich features for perceptual quality assessment of UGC videos.
\newblock In \emph{CVPR}, 13435--13444.

\bibitem[{Wen et~al.(2024)Wen, Li, Zhang, Liao, Li, Zhang, and Ma}]{paper106}
Wen, W.; Li, M.; Zhang, Y.; Liao, Y.; Li, J.; Zhang, L.; and Ma, K. 2024.
\newblock Modular blind video quality assessment.
\newblock In \emph{CVPR}, 2763--2772.

\bibitem[{Wu et~al.(2022)Wu, Chen, Hou, Liao, Wang, Sun, Yan, and
  Lin}]{paper29}
Wu, H.; Chen, C.; Hou, J.; Liao, L.; Wang, A.; Sun, W.; Yan, Q.; and Lin, W.
  2022.
\newblock Fast-vqa: Efficient end-to-end video quality assessment with fragment
  sampling.
\newblock In \emph{ECCV}, 538--554.

\bibitem[{Wu et~al.(2023{\natexlab{a}})Wu, Chen, Liao, Hou, Sun, Yan, Gu, and
  Lin}]{paper19}
Wu, H.; Chen, C.; Liao, L.; Hou, J.; Sun, W.; Yan, Q.; Gu, J.; and Lin, W.
  2023{\natexlab{a}}.
\newblock Neighbourhood representative sampling for efficient end-to-end video
  quality assessment.
\newblock \emph{IEEE TPAMI}.

\bibitem[{Wu et~al.(2023{\natexlab{b}})Wu, Chen, Liao, Hou, Sun, Yan, and
  Lin}]{paper23}
Wu, H.; Chen, C.; Liao, L.; Hou, J.; Sun, W.; Yan, Q.; and Lin, W.
  2023{\natexlab{b}}.
\newblock Discovqa: Temporal distortion-content transformers for video quality
  assessment.
\newblock \emph{IEEE TCSVT}.

\bibitem[{Wu et~al.(2023{\natexlab{c}})Wu, Zhang, Liao, Chen, Hou, Wang, Sun,
  Yan, and Lin}]{paper64}
Wu, H.; Zhang, E.; Liao, L.; Chen, C.; Hou, J.; Wang, A.; Sun, W.; Yan, Q.; and
  Lin, W. 2023{\natexlab{c}}.
\newblock Exploring video quality assessment on user generated contents from
  aesthetic and technical perspectives.
\newblock In \emph{ICCV}, 20144--20154.

\bibitem[{Wu et~al.(2023{\natexlab{d}})Wu, Zhang, Liao, Chen, Hou, Wang, Sun,
  Yan, and Lin}]{paper63}
Wu, H.; Zhang, E.; Liao, L.; Chen, C.; Hou, J.; Wang, A.; Sun, W.; Yan, Q.; and
  Lin, W. 2023{\natexlab{d}}.
\newblock Towards Explainable In-the-Wild Video Quality Assessment: A Database
  and a Language-Prompted Approach.
\newblock In \emph{ACM MM}, 1045–1054.

\bibitem[{Wu et~al.(2024)Wu, Zhang, Zhang, Chen, Liao, Li, Gao, Wang, Zhang,
  Sun et~al.}]{paper105}
Wu, H.; Zhang, Z.; Zhang, W.; Chen, C.; Liao, L.; Li, C.; Gao, Y.; Wang, A.;
  Zhang, E.; Sun, W.; et~al. 2024.
\newblock Q-ALIGN: teaching LMMs for visual scoring via discrete text-defined
  levels.
\newblock In \emph{ICML}, 54015--54029.

\bibitem[{Wu et~al.(2023{\natexlab{e}})Wu, Hu, Xiao, Deng, Li, Chen, and
  Li}]{paper131}
Wu, W.; Hu, S.; Xiao, P.; Deng, S.; Li, Y.; Chen, Y.; and Li, K.
  2023{\natexlab{e}}.
\newblock Video quality assessment based on swin transformer with
  spatio-temporal feature fusion and data augmentation.
\newblock In \emph{CVPR}, 1846--1854.

\bibitem[{Xing et~al.(2024)Xing, Li, Wang, Zhu, and Cao}]{paper124}
Xing, F.; Li, M.; Wang, Y.-G.; Zhu, G.; and Cao, X. 2024.
\newblock Clipvqa: Video quality assessment via clip.
\newblock \emph{IEEE Trans. on Broadcast.}

\bibitem[{Xu et~al.(2021)Xu, Ghosh, Huang, Okhonko, Aghajanyan, Metze,
  Zettlemoyer, and Feichtenhofer}]{paper127}
Xu, H.; Ghosh, G.; Huang, P.-Y.; Okhonko, D.; Aghajanyan, A.; Metze, F.;
  Zettlemoyer, L.; and Feichtenhofer, C. 2021.
\newblock VideoCLIP: Contrastive Pre-training for Zero-shot Video-Text
  Understanding.
\newblock In \emph{EMNLP}, 6787--6800.

\bibitem[{Xu et~al.(2024)Xu, Xie, Tan, Huang, Howes, Sharma, Li, Ghosh,
  Zettlemoyer, and Feichtenhofer}]{paper112}
Xu, H.; Xie, S.; Tan, X.~E.; Huang, P.~Y.; Howes, R.; Sharma, V.; Li, S.~W.;
  Ghosh, G.; Zettlemoyer, L.; and Feichtenhofer, C. 2024.
\newblock DEMYSTIFYING CLIP DATA.
\newblock In \emph{ICLR}.

\bibitem[{Xu et~al.(2014)Xu, Ye, Liu, and Doermann}]{paper33}
Xu, J.; Ye, P.; Liu, Y.; and Doermann, D. 2014.
\newblock No-reference video quality assessment via feature learning.
\newblock In \emph{ICIP}, 491--495.

\bibitem[{Yang et~al.(2024)Yang, Zhang, Wang, and Xie}]{paper122}
Yang, L.; Zhang, R.-Y.; Wang, Y.; and Xie, X. 2024.
\newblock Mma: Multi-modal adapter for vision-language models.
\newblock In \emph{CVPR}, 23826--23837.

\bibitem[{Ying et~al.(2021)Ying, Mandal, Ghadiyaram, and Bovik}]{paper16}
Ying, Z.; Mandal, M.; Ghadiyaram, D.; and Bovik, A. 2021.
\newblock Patch-VQ:'Patching Up'the video quality problem.
\newblock In \emph{CVPR}, 14019--14029.

\bibitem[{You et~al.(2025)You, Cai, Gu, Xue, and Dong}]{paper123}
You, Z.; Cai, X.; Gu, J.; Xue, T.; and Dong, C. 2025.
\newblock Teaching large language models to regress accurate image quality
  scores using score distribution.
\newblock In \emph{CVPR}, 14483--14494.

\bibitem[{Yuan et~al.(2024)Yuan, Liu, Li, Sun, Sun, Gong, Hao, Zhou, and
  Tang}]{paper108}
Yuan, K.; Liu, H.; Li, M.; Sun, M.; Sun, M.; Gong, J.; Hao, J.; Zhou, C.; and
  Tang, Y. 2024.
\newblock PTM-VQA: efficient video quality assessment leveraging diverse
  pretrained models from the wild.
\newblock In \emph{CVPR}, 2835--2845.

\bibitem[{Zhai et~al.(2023)Zhai, Mustafa, Kolesnikov, and Beyer}]{paper110}
Zhai, X.; Mustafa, B.; Kolesnikov, A.; and Beyer, L. 2023.
\newblock Sigmoid loss for language image pre-training.
\newblock In \emph{ICCV}, 11975--11986.

\bibitem[{Zhang et~al.(2024)Zhang, Huang, Jin, and Lu}]{paper114}
Zhang, J.; Huang, J.; Jin, S.; and Lu, S. 2024.
\newblock Vision-language models for vision tasks: A survey.
\newblock \emph{IEEE TPAMI}, 46(8): 5625--5644.

\bibitem[{Zhang et~al.(2021)Zhang, Fang, Zhang, Gao, Li, Dai, Qiao, and
  Li}]{paper126}
Zhang, R.; Fang, R.; Zhang, W.; Gao, P.; Li, K.; Dai, J.; Qiao, Y.; and Li, H.
  2021.
\newblock Tip-adapter: Training-free clip-adapter for better vision-language
  modeling.
\newblock \emph{arXiv preprint arXiv:2111.03930}.

\bibitem[{Zhang et~al.(2023{\natexlab{a}})Zhang, HaoChen, Huang, Wang, Zou, and
  Yeung}]{paper117}
Zhang, Y.; HaoChen, J.~Z.; Huang, S.-C.; Wang, K.-C.; Zou, J.; and Yeung, S.
  2023{\natexlab{a}}.
\newblock Diagnosing and Rectifying Vision Models using Language.
\newblock In \emph{ICLR}.

\bibitem[{Zhang et~al.(2023{\natexlab{b}})Zhang, Wu, Sun, Tu, Lu, Min, Chen,
  and Zhai}]{paper43}
Zhang, Z.; Wu, W.; Sun, W.; Tu, D.; Lu, W.; Min, X.; Chen, Y.; and Zhai, G.
  2023{\natexlab{b}}.
\newblock MD-VQA: Multi-dimensional quality assessment for UGC live videos.
\newblock In \emph{CVPR}, 1746--1755.

\bibitem[{Zhao et~al.(2023)Zhao, Yuan, Sun, and Wen}]{paper45}
Zhao, K.; Yuan, K.; Sun, M.; and Wen, X. 2023.
\newblock Zoom-VQA: Patches, Frames and Clips Integration for Video Quality
  Assessment.
\newblock In \emph{CVPR}, 1302--1310.

\bibitem[{Zhou et~al.(2022{\natexlab{a}})Zhou, Yang, Loy, and Liu}]{paper119}
Zhou, K.; Yang, J.; Loy, C.~C.; and Liu, Z. 2022{\natexlab{a}}.
\newblock Conditional prompt learning for vision-language models.
\newblock In \emph{CVPR}, 16816--16825.

\bibitem[{Zhou et~al.(2022{\natexlab{b}})Zhou, Yang, Loy, and Liu}]{paper118}
Zhou, K.; Yang, J.; Loy, C.~C.; and Liu, Z. 2022{\natexlab{b}}.
\newblock Learning to prompt for vision-language models.
\newblock \emph{IJCV}, 130(9): 2337--2348.

\end{thebibliography}

\end{document}